\documentclass[journal]{IEEEtran}
\usepackage[table,xcdraw]{xcolor}
\usepackage{subfigure}
\usepackage{graphicx}
\usepackage{stfloats}
\usepackage{color}
\usepackage{float}
\usepackage{amsmath}
\usepackage{amssymb}
\usepackage{bm}
\usepackage{epstopdf}
\usepackage[noend]{algpseudocode}
\usepackage{algorithmicx,algorithm}

\ifCLASSINFOpdf
\else
\fi

\begin{document}
%
\title{\emph{cm}SalGAN: RGB-D Salient Object Detection with Cross-View Generative Adversarial Networks}
%
%
%


%

\author{Bo Jiang, 
        Zitai Zhou, 
        Xiao Wang,
        Jin Tang
         and Bin Luo 
\IEEEcompsocitemizethanks{
\IEEEcompsocthanksitem
This work is supported in part by Major Project for New Generation of AI under Grant (No. 2018AAA0100400); NSFC Key
Projects of International (Regional) Cooperation and Exchanges under Grant (61860206004); Open fund for Discipline Construction, Institute of Physical Science and Information Technology, Anhui University
\IEEEcompsocthanksitem
Bo Jiang is with Key Laboratory of Intelligent Computing and Signal Processing of Ministry of Education, School of
Computer Science and Technology, Anhui University Institute of Physical Science and Information Technology, Anhui University
\IEEEcompsocthanksitem Zitai Zhou, Xiao Wang and Bin Luo are
with Key Laboratory of Intelligent Computing and Signal Processing of Ministry of Education, School of Computer Science and Technology, Anhui University, Hefei 230601, China. 
\IEEEcompsocthanksitem Jin Tang is with Anhui Provincial Key Laboratory of Multimodal Cognitive Computation, School of Computer Science and Technology, Anhui University, 230601, China\protect\\
E-mail: \{jiangbo, tj, luobin\}@ahu.edu.cn, zicair@sina.com, wangxiaocvpr@foxmail.com.
}
\thanks{Corresponding author: Xiao Wang}}




\maketitle

\begin{abstract}
Image salient object detection (SOD) is an active research topic in computer vision and multimedia area. Fusing complementary information of RGB and depth has been demonstrated to be effective for image salient object detection which is known as RGB-D salient object detection problem. The main challenge for RGB-D salient object detection is how to exploit the salient cues of both intra-modality (RGB, depth) and cross-modality simultaneously which is known as cross-modality detection problem. In this paper, we tackle this challenge by designing a novel cross-modality Saliency Generative Adversarial Network (\emph{cm}SalGAN). \emph{cm}SalGAN aims to learn an optimal view-invariant and consistent pixel-level representation for RGB and depth images  via a novel adversarial learning framework, which thus incorporates both information of intra-view and correlation information of cross-view images simultaneously for RGB-D saliency detection problem.
To further improve the detection results, the attention mechanism and edge detection module are also incorporated into \emph{cm}SalGAN.
The entire \emph{cm}SalGAN  can be trained in an end-to-end manner by using the standard deep neural network framework. Experimental results show that \emph{cm}SalGAN  achieves the new state-of-the-art RGB-D saliency detection performance on several benchmark datasets.
\end{abstract}

\begin{IEEEkeywords}
RGB-D Saliency Detection, Generative Adversarial Learning, Multi-view Learning
\end{IEEEkeywords}

\IEEEpeerreviewmaketitle

\section{Introduction}

As an important research topic in computer vision and multimedia area, salient object detection (SOD) has attracted more and more attention in recent years. It aims at highlighting salient object regions from the given image and has been widely used in object-level applications in different fields, such as image understanding, object detection, and tracking. The main issues for the SOD task are twofold, i.e., 1) pixel-level representation and 2) saliency prediction/estimation. In the early years, many traditional methods have been proposed for saliency detection by exploiting some low-level feature representations, such as color, HOG, etc. In recent years, with the development of deep learning-based  representation methods, salient object detection has been significantly improved via CNN based pixel-level representation. However, although the salient object detection has made great progress in recent years \cite{jerripothula2016image, wang2019learning, perazzi2012saliency, cong2016saliency, cheng2014depth, niu2012leveraging, wang2019quality, rosani2015eventmask, park2017transfer, tang2019salient, yu2015computational, liang2016looking, peng2019automatic, xu2019video, li2019accurate, deng2019saliency}, it is still a challenging problem mainly due to the complicated background and different lighting conditions in the images.

\begin{figure}
\centering
\includegraphics[width=0.9\columnwidth]{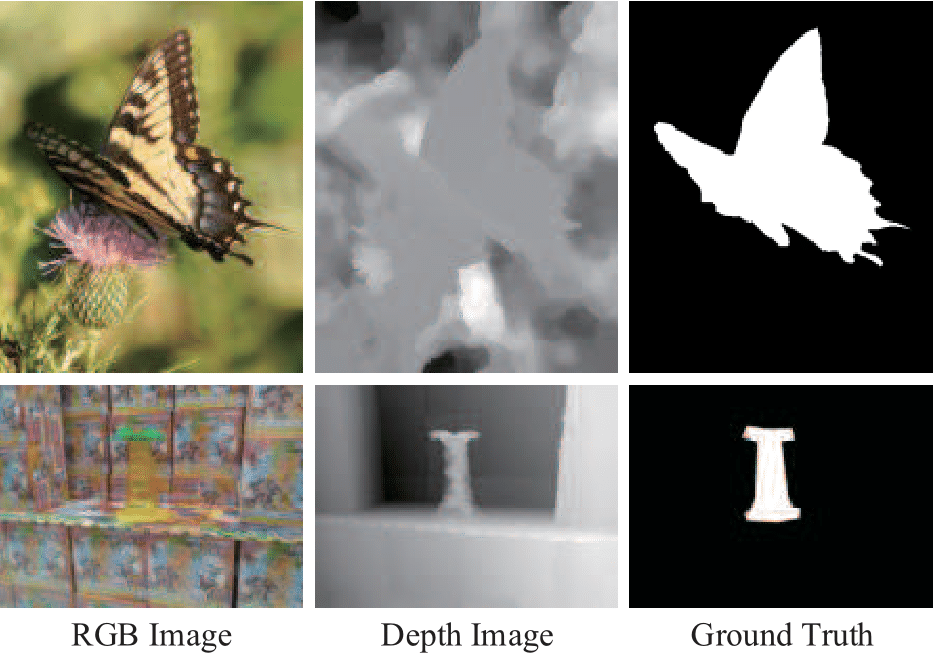}
\caption{The illustration of RGB-D saliency detection. The RGB and Depth information can complementary to each other.  }
\label{fig:example}
\end{figure}

Recently, many works attempt to utilize more modalities to boost the performance of SOD results.
One popular way is to integrate the depth information into RGB saliency detection, which is known as
RGB-D saliency detection problem.
The key idea of RGB-D saliency detection problem is to exploit the complementary information from given modalities for better saliency detection. Since the RGB image contains rich color and texture information while the depth image contains rich depth and contour information. Therefore, how to fuse these modalities in an adaptive and complementary manner is the main challenge of this task. As shown in Figure \ref{fig:example}, the `butterfly' object in   RGB image is appeared  better than that in the depth image, while the salient object in depth image shows better quality in the second row of Figure \ref{fig:example}. Therefore, how to fuse these modalities (RGB, depth) adaptively is the key issue to the success of RGB-D salient object detection.

Most of previous works handle the multi-modal fusion problem by \emph{either} serializing the RGB-D channels directly for data representation~\cite{chen2018progressively,chen2019multi,zhao2019contrast,chen2019three,piao2019depth} \emph{or} processing the representation of each modality independently and then combining them together for the final multi-modal representation~\cite{wang2019quality,han2017cnns,shigematsu2017learning,wang2019adaptive,fan2019rethinking,LIU2020210}. Although these strategies can obtain encouraging results, they are still difficult to fully explore cross-modal complementarity.

In this paper, we propose to address this issue by adopting a generative adversarial learning framework.
We develop a novel cross-modality Saliency Generative Adversarial Network (\emph{cm}SalGAN) for RGB-D salient object detection.
Overall, \emph{cm}SalGAN aims to learn an optimal modality-invariant and thus can fuse pixel-level representation of RGB and depth images together via a novel adversarial learning framework, which thus incorporates both information of intra-modality and correlation
information of cross-modality images simultaneously for the RGB-D saliency detection problem.
Specifically speaking, we use two encoder-decoder networks to extract the pixel-level features of RGB and depth images, respectively.
We design a novel cross-modality adversarial learning mechanism to boost the representation learning of different modalities, which can achieve the purpose of information fusion.
To further improve the detection results, the attention mechanism and edge detection module are also incorporated into \emph{cm}SalGAN.
The entire \emph{cm}SalGAN  can be trained in an end-to-end manner by using both saliency prediction cross-entropy loss and cross-modality adversarial learning loss.

Note that, generative adversarial networks (GANs) have been designed for image saliency detection tasks ~\cite{pan2017salgan,wang2019quality,LIU2020210} and they focus on adversarial learning between saliency prediction and ground-truth salient object.
In contrast to previous works, \emph{cm}SalGAN aims to conduct adversarial learning between different modality representations. That is, in \emph{cm}SalGAN the generator and discriminator beat each other as a minimax game to learn discriminative common representation of heterogeneous multi-modality data for the final saliency prediction.

Overall, the main contributions of this paper are summarized as the following three aspects:
\begin{itemize}
\item We propose to tackle the problem of cross-modality RGB-D image representation for saliency estimation
by exploiting the generative adversarial representation learning. To the best of our knowledge, it is the first work to conduct multi-modality adversarial learning for RGB-D image representation.

\item A loss function is designed for cross-modality generative adversarial network (\emph{cm}SalGAN) training to learn a discriminative common presentation for each pixel of RGB and depth images.

\item Comprehensive experiments on several widely used RGB-D benchmark datasets validate the effectiveness of the proposed \emph{cm}SalGAN approach. It is also worthy to note that \emph{cm}SalGAN achieves the new state-of-the-art performance on these benchmarks.
\end{itemize}

The remainder of this paper is organized as follows.
In section II, we briefly review some related works on RGB-D saliency detection and Generative Adversarial Networks (GANs).
We present the detail of \emph{cm}SalGAN in section III.
In section IV, we implement \emph{cm}SalGAN on several benchmarks to demonstrate the effectiveness of the proposed model.

\section{Related Work}

In this work, we briefly review the related papers on RGB-D saliency detection and generative adversarial networks.

\subsection{RGB-D saliency detection}
Most of previous works focus on detect saliency objects on the RGB images, for example, Peng \emph{et al.} \cite{peng2019automatic} propose an approach to extract the salient objects in videos automatically. Specifically, they utilize the weighted multiple manifold ranking algorithm to identify object-like regions in each frame. Then,
they compute motion cues to estimate the motion saliency and localization prior and finally estimate the superpiexel-level object labelling across all frames with a new energy function.
Wang \emph{et al.} \cite{wang2019learning} propose a learning criterion requires the output of the model to be close to the original output. They also propose a hierarchical attribution fusion scheme to enhance the smooth the optimized saliency masks. Koteswar \emph{et al.} \cite{jerripothula2016image} adopt the co-saliency detection technique to help exploit the inter-image information, then they perform single-image segmentation on each individual image.

Recently, many works have been proposed for the RGB-D saliency detection problem.
The core aspect of these works is how to exploit the salient cues of both inter-modality and intra-modality for the final saliency prediction.

One kind of popular way is to first serialize the RGB-D channels directly for image representation and then conduct
saliency prediction. For example, Chen \emph{et al}.~\cite{chen2018progressively} propose a PCA network for RGB-D saliency detection which uses a novel complementarity-aware fusion module to deal with the complementarity of two modal information.
In work~\cite{chen2019multi}, the authors propose a multi-modal fusion network with Multi-scale Multi-path and Cross-modal Interactions (MMCI) network for RGB-D saliency detection.
The method aims to use a multi-scale multi-path manner to diversify the contributions of each modality by using a cross-modal interaction.
Zhao~\emph{et al}. \cite{zhao2019contrast} propose a network named Contrast Prior and Fluid Pyramid integration (CPFP) for RGB-D saliency detection which
integrates multi-scale cross-modal features by using a pyramid integration model.
Chen \emph{et al}. \cite{chen2019three} recently propose Three-stream Attention-aware Network (TANet) for RGB-D saliency detection by using a novel triplet-stream multi-modal fusion architecture to extract cross-modal complementary features. Piao \emph{et al}. \cite{piao2019depth} fuse the cross-modal features and then apply a recurrent attention module to boost the performance.

Another way to handle the multi-modal fusion problem is first processing the representation of each modality independently and then combining them together for the final multi-modal representation.
For example, In CTMF~\cite{han2017cnns}, it first uses a two-stream architecture to exploit the multi-modal features for RGB and depth images respectively.
Then, it aims to {merge} the representation of the two views to obtain the final saliency maps by using a multi-view CNN fusion model.
In work \cite{shigematsu2017learning}, it first extracts handcrafted RGB and depth features in a two-stream network
and then fuse them together for RGB-D saliency detection.
Wang \emph{et al}. \cite{wang2019adaptive} recently propose to employ the U-Net \cite{ronneberger2015u} framework to learn a switch map to estimate the weights for fusing RGB and depth saliency maps together. In addition, inspired by salGAN \cite{pan2017salgan}, Wang \emph{et al}. \cite{wang2019quality} adopt Generative adversarial networks (GANs)
for RGB-D saliency detection.
They first use MSE and adversarial loss function to extract salient cues for  RGB and depth modality, respectively. Then, they employ a reinforcement learning architecture to adaptively fuse these cues together for final saliency prediction. Fan \emph{et al}. \cite{fan2019rethinking} propose a depth depurator unit (DDU) to filter out the low-quality depth map and then fuse the cross-modal feature for learning. Liu \emph{et al}. \cite{LIU2020210} utilize double-stream encoder-decoder network to extract the cross-modal feature and then propose a gated fusion module and employ adversarial learning for RGB-D saliency detection.

Different from previous related works~\cite{pan2017salgan,wang2019quality,LIU2020210},
the proposed \emph{cm}SalGAN aims to conduct adversarial learning between different modality representations and {to learn} a kind of discriminative common representation for both RGB and depth data for saliency prediction.

\subsection{Generative adversarial networks}

Generative adversarial networks (GANs) \cite{goodfellow2014generative} was
originally proposed by Goodfellow \emph{et al}. and have received increasing attention
in the fields of machine learning and computer vision fields.
Recently, GANs have been exploited for cross-modality visual data representation~\cite{bousmalis2017unsupervised,dai2018cross,lekic2019automotive,gammulle2019coupled,guo2019fusegan,yu2019ea,xie2020multi}.
For example, Dai \emph{et al}. \cite{dai2018cross} propose a \emph{cm}GAN network for cross-modality Re-ID task which
uses GANs to learn feature representation from different modalities.
Lekic \emph{et al}. \cite{lekic2019automotive} employs GANs to fuse the radar sensor measurements with the camera images.
Gammulle \emph{et al}. \cite{gammulle2019coupled} apply GANs  for fine-grained human action segmentation.
Li \emph{et al}. \cite{li2018self} employ GANs to further enhance the retrieval accuracy.
Similarly, Zhang \emph{et al}. \cite{zhang2018sch} propose SCH-GAN for semi-supervised cross-modal hashing representation.
Dou \emph{et al}. \cite{dou2018unsupervised} design a cross-modal biomedical image segmentation network via an adversarial learning.
Wang \emph{et al}. \cite{wang2017adversarial} use GANs to seek an common subspace for cross-modal retrieval.
Similarly, Peng \emph{et al}. \cite{peng2019cm} use GANs to exploit the cross-modal common representations.
Ma \emph{et al}. \cite{ma2019fusiongan} adopt GANs to fuse the information of visible and infrared images.
Zhao \emph{et al}. \cite{zhao2019simultaneous} propose a color-depth conditional GAN to concurrently resolve the problem of depth super-resolution and color super-resolution in 3D videos.

Recently, some works also employ  GANs for saliency detection tasks. This is because the \emph{pixel-level} measure of saliency results, such as binary cross-entropy loss, can be designed for per-pixel category prediction~\cite{luo2018macroAdversarial}. However, this pixel-level model generally penalizes the false prediction on every pixel which thus lacks of explicitly modeling the correlation among adjacent pixels and may lead to local inconsistency and semantic inconsistency in the global saliency map prediction. Therefore, some researchers attempt to introduce some \emph{high-level} evaluation criteria, such as adversarial network, to handle these issues.
For example, previous works proposed in~\cite{pan2017salgan,wang2019GANTrack, luc2016semantic} and \cite{wang2019quality} all adopt adversarial learning mechanism and achieve better results on RGB or RGB-D related tasks. The adversarial learning mechanism judges whether a given saliency result is real or fake by the joint configuration of many label variables, and thus can enforce high-level consistency.
Specifically, Fernando \emph{et al}.~\cite{fernando2018task} apply GANs for human saliency estimation to jointly model the contextual semantic and relations in different tasks.  Pan \emph{et al.} \cite{pan2017salgan} propose Saliency GAN (SalGAN) for saliency prediction task which is trained with MSE and adversarial loss functions.  SalGAN360 \cite{chao2018salgan360} further extends this framework for the  $360^\circ$ image-based saliency prediction. Wang \emph{et al}. \cite{wang2019quality} jointly use MSE and adversarial loss function to predict the results of two modalities and then use reinforcement learning to learn the weighted value of the two results. Liu \emph{et al} \cite{LIU2020210}. propose a gated fusion module for adversarial learning in which the purpose of the discriminator is to learn the gated fusion weights for RGB-D feature extraction.

Previous works~\cite{pan2017salgan,wang2019quality,LIU2020210} focus on adversarial learning between saliency prediction and ground-truth salient object in the later stage. Different from previous adversarial learning based saliency estimation methods, we tackle the problem of cross-modality RGB-D image representation for saliency estimation by exploiting the generative adversarial representation learning. Specifically, \emph{cm}SalGAN aims to conduct adversarial learning between different modality representations and to make the feature of the two modalities complement each other more effectively.
To our best knowledge, it is the first work to conduct multi-modality adversarial learning for RGB-D image representation and saliency detection problem, although multi-modality adversarial learning has been studied in other tasks, as summarized in before.

\section{The Proposed Method}

\begin{figure*}[htb]
\centering
\includegraphics[width=0.95\textwidth]{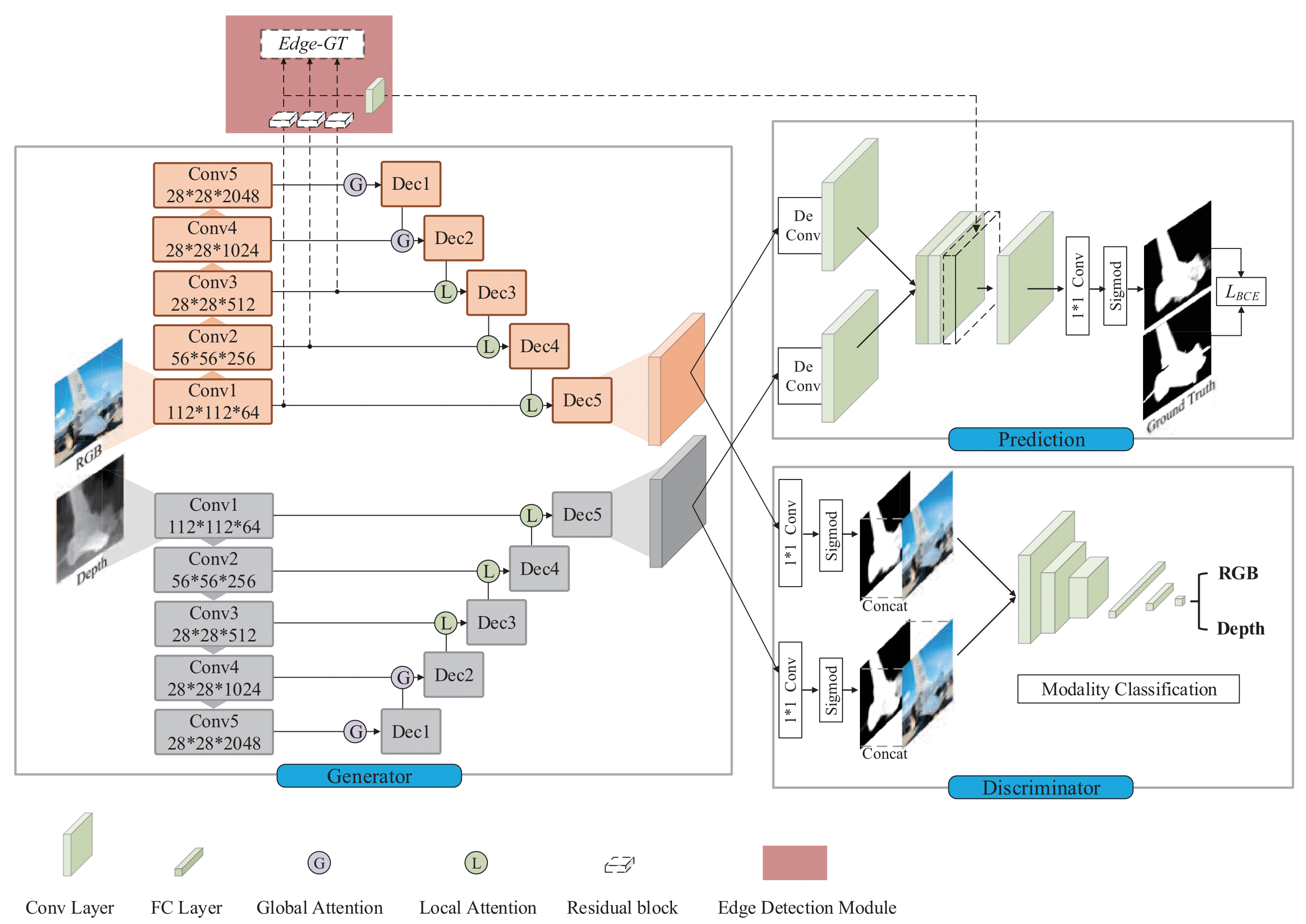}
\caption{The illustration of the proposed \emph{cm}SalGAN network for salient object detection. Edge-GT represents the ground truth of the input image when training the edge detection module.}
\label{fig:network}
\end{figure*}

In this section, we  describe the detail of our \emph{cm}SalGAN network.
Figure 2 shows the overall of the proposed \emph{cm}SalGAN network which mainly consists of the following four components:

\begin{itemize}
\item {\bfseries Two-stream Generator}:
In this paper, we adopt a two-stream network as the generator to learn the feature representation for RGB and depth images respectively. This generator contains two encoder-decoder networks which share the initial weights and are  trained to learn their respective network weight parameters.

\item {\bfseries Adversarial Feature Learning}: Following the framework of generative adversarial learning, we introduce a discriminator, \emph{i.e.} Adversarial Feature Learning (AFL) module, to make the features from the two branches combat to each other. This module will enhance the consistent constraint of our saliency detection framework from the perspective of feature learning.

\item {\bfseries Edge Detection Module}: Inspired by existing works \cite{xie2015holistically, chen2016semantic} which adopt additional edge information for fine-grained segmentation, we also utilize the edge features for more accurate saliency detection in our approach.

\item {\bfseries Saliency Prediction}:
We use the de-convolutional layer to restore the resolution of two modalities and feed them into the convolutional layer and ReLU layer to learn the parameters for adaptive fusion. Then, we employ a sigmoid activation function to generate the final saliency prediction.
\end{itemize}

In the following subsections, we will present the details of each component as mentioned above.

\subsection{Two-stream Generator}

In the feature extraction phase, we utilize a two-stream generator that contains two encode-decoder networks to learn the deep representation of RGB and depth images, respectively. Specifically, the encoder module is a truncated ResNet-50 \cite{he2016deep} network (fully connected layers removed) with hole algorithm \cite{chen2017deeplab} which can keep the resolutions of feature maps unchanged. The weights of the encoder {are} initialized with a pre-trained model on the ImageNet dataset~\cite{deng2009imagenet} for object classification. The input samples are all resized into $224\times224$ and the output feature map $\textit{Enc}^{i}$ from the 1st to 5th convolutional layer ($Conv_{i}, i\in \{1,2,3,4,5\}$) has the resolution of $112, 56, 28, 28$ and $28$ respectively. The decoder contains deconvolutional layers which are used to increase the resolution of the encoded feature map. We use $Decoder_{i}, i\in \{1,2,3,4,5\}$ to denote the corresponding decoder layers. It is worthy to note that the input of depth image is preprocessed into three channels to make it consistent with the RGB branch before feed into the corresponding encoder network.

To obtain better feature representation, we follow the idea of U-Net~\cite{ronneberger2015u} to
fully utilize both low-level and high-level feature maps in the decoder network by using some skip network connections.
This is because different layers contain different information. For example, the lower layers of encoder generally involve rich details while the high-level layers usually contain more semantic information.
Therefore, we can obtain a kind of richer decoding feature map $\textit{Dec}^{i}$ by fusing encoder's feature map $\textit{Enc}^{i}$ and decoder's feature map $\textit{Dec}^{i-1}$ together.
In addition, inspired by work~\cite{liu2018picanet}, we employ both local and global attention schemes in the decoder network to obtain better feature representation. Specifically, in the global part, we use LSTM \cite{graves2013hybrid} in $Decoder_{1}$ and $Decoder_{2}$ layers to obtain global context information by scanning the input feature maps along both horizontal and vertical directions. In global and local decoder layers, we use an attention mechanism to incorporate multi-scale context information.
Given a convolutional feature map $F\in\mathbb{R}^{W\times H\times C}$ where $W$, $H$ and $C$ denote its width, height and number of channels respectively, we first use a convolutional layer (kernel size is $1\times 1$) to transform it into a feature map with dimension $K=W\times H$. Then, we extract the feature vector $\mathbf{x}^{w, h}=(x^{w, h}_1, x^{w, h}_2, \cdots, x^{w, h}_{K})$ from each spatial location ($w, h$) and the dimension of this vector is $K$. Similar to previous work~\cite{liu2018picanet}, we adopt Softmax function to normalize vector $\mathbf{x}^{w, h}$ to obtain attention weights $\bm{\alpha}^{w,h}$, the $\bm{\alpha}^{w,h}$ can be unfold to $\bm{\alpha}^{w,h}=(\alpha^{w, h}_1, \alpha^{w, h}_2, \cdots, \alpha^{w, h}_{K})$.
Using the learned $\bm{\alpha}^{w,h}$, we can obtain the attended feature $F_{attention}^{w,h}$ in each spatial location ($w, h$) as~\cite{liu2018picanet}
\begin{equation}
F_{attention}^{w,h}=\sum_{i=1}^{K}\alpha _{i}^{w,h} f_{i}
\end{equation}
where $f_{i}\in \mathbb{R}^{C}$ represents the feature vector at spatial location ($w, h$) in $F$. Thus we can get the final attention weighted feature $F_{attention}$.

\subsection{Adversarial Feature Learning}

In this paper, we employ an adversarial learning to further explore cross-modal complementarity for RGB-D saliency detection task.
Generally speaking, we introduce the idea of multi-view learning into the framework of generative adversarial networks~\cite{dai2018cross}. More specifically, we take the output features of RGB and depth modality from the generator as an input of the convolutional (kernel size is $1\times 1$) and sigmoid layer and generate saliency results from both modalities respectively. Then, we concatenate the predicted saliency results with RGB image and feed them into the discriminator to judge the given results belong to RGB or depth modality. Through the adversarial multi-view learning, we can train the generator to learn a consistent feature representation of the RGB-D image.

\begin{table}[t]
\center
\caption{The detailed architecture of discriminator D.}\label{configurationDiscriminator}
\begin{tabular}{l|cc|c}
\hline
\hline
layer   &kernel  &activation  &out-channel\\
\hline
conv1	&1$\times$1		&ReLU   &3\\
conv2	&3$\times$3		&ReLU	&32\\
max-Pooling &2$\times$2	&-      &32\\
\hline
conv3	 &3$\times$3	&ReLU   &64\\
conv4	 &3$\times$3	&ReLU	&64\\
max-Pooling	&2$\times$2	&-      &64\\
\hline
conv5	 &3$\times$3	&ReLU   &64\\
conv6	 &3$\times$3	&ReLU	&64\\
max-Pooling	&2$\times$2 &-     &64\\
\hline
fc7	  &-	&tanh       &100\\
fc8	  &-	&tanh	    &2\\
fc9	  &-	&sigmoid  	&1\\
\hline
\hline
\end{tabular}
\end{table}

The discriminator $D$ used in our network is a standard convolutional network that contains convolutional layer, ReLU, max-pooling, fully connected layer and sigmoid activation layer. Our discriminator is designed based on \cite{pan2017salgan} and the detailed architecture can be found in Table \ref{configurationDiscriminator}. The discriminator is used to judge the given input belongs to the RGB modality or depth modality which will be beneficial for learning {a discriminative} feature representation. Concretely, after obtaining the feature maps $\textit{Dec}_{rgb}^{5}$ and $\textit{Dec}_{depth}^{5}$ from the two-stream network, we first employ a $1\times1$ convolutional operation on these feature maps and use a sigmoid activation function to obtain the final saliency map $\textit{S}_{r}$ and $\textit{S}_{d}$ whose resolution are all $112\times 112\times 1$. To make the predicted saliency result consistent with the original image on the resolution,  we conduct bilinear interpolation on the saliency map $\textit{S}_{r}$ and $\textit{S}_{d}$ and concatenate them with RGB image to form 4-channel feature maps, respectively. Finally, the feature maps are fed into the adversarial feature learning (AFL) module to achieve adversarial multi-view learning across different modalities.
During the training of our AFL module in which generator and discriminator compete with each other in the form of \emph{mini-max} game to learn the common representation. Similar to \cite{pan2017salgan}, the loss function of discriminator can be written as:
\begin{equation}
\label{ganLoss}
\mathcal{L}_{\mathcal{D}}=\mathcal{L}\left ( D\left ( I_r, S_{r} \right ),1 \right )+\mathcal{L}\left ( D\left ( I_r, S_{d} \right ),0 \right )
\end{equation}
where $\mathcal{L}$ denotes binary cross entropy loss, and $D(\cdot,\cdot)$ is the discriminator function used in the adversarial learning procedure.
$I_r$ represents the corresponding original input RGB image. Here we use 1 to denote the target category of RGB sample and 0 for depth sample.

Therefore, the final loss function of the proposed algorithm is formulated as
\begin{equation}
\mathcal{L}=\mathcal{L}_{BCE} + \mathcal{L}\left ( D\left ( I_r, S_{r} \right ),0 \right ) + \mathcal{L}\left ( D\left ( I_r, S_{d} \right ),1 \right )
\end{equation}
where $\mathcal{L}_{BCE}$~\cite{de2005tutorial} is defined as
\begin{equation}
\small
\label{bceloss}
\mathcal{L}_{\mathrm{BCE}}= \frac{1}{{W\times H}} \sum_{i=1}^{W}\sum_{j=1}^{H}[( 1-S_{ij}  )\log   ( 1-\hat{S}_{ij} )-S_{ij}\log  ( \hat{S}_{ij} )]
\end{equation}
where $\hat{S}_{ij}, S_{ij}$ represent the saliency map and corresponding ground truth, respectively.

\subsection{Edge Detection Module}

In order to estimate  the final salient object more accurately, inspired by recent work~\cite{liu2019simple}, we further introduce an additional edge detection module on the basis of existing network to extract edge features and fuse them into our saliency prediction.
Different from previous work~\cite{liu2019simple}, we integrate the residual convolutional blocks into the first three convolutional blocks of the encoder to implement feature transformation and edge feature encoding.
Since the first three convolutional blocks of the encoder contain more detailed information and {they} will be more desirable to extract the edge information. Through the first three residual convolutional blocks, we can obtain three kinds of features and all of which have $16$ channels. These features are concatenated together and fed into a $1\times 1$ convolutional layer to generate the feature map whose dimension is $64$.
The parameters of the generator are fixed when training the edge detection module and saliency prediction module.

\subsection{Saliency Prediction}
After we obtain the convolutional features of two modalities, we upsample $\textit{Dec}_{rgb}^{5}$ and $\textit{Dec}_{depth}^{5}$ to make them have the same resolution via deconvolutional layers. Then, these two feature maps are concatenated together and fed into a convolutional layer and ReLU layer.
Formally, the fused feature map can be transformed into saliency results via a $1\times 1$ convolutional operation and sigmoid layer. In this paper, we adopt the commonly used binary cross-entropy (BCE) loss~\cite{de2005tutorial} to measure the distance between our saliency prediction and the ground truth saliency map.
The loss function $\mathcal{L}_{\mathrm{BCE}}$ is {defined as Eq. (\ref{bceloss})},
where $\hat{S}_{ij}, S_{ij}$ represent the saliency map and corresponding ground truth, respectively.

\section{Experiments}

To evaluate the effectiveness of the proposed \emph{cm}SalGAN approach,
we test it on three benchmark datasets.
In the following, we first introduce the datasets and evaluation metrics used in our experiments.
Then, we present the implementation details of our \emph{cm}SalGAN  saliency detection algorithm.
Finally, we compare our method with other state-of-the-art RGB-D saliency detection algorithms and further conduct the ablation studies for the proposed \emph{cm}SalGAN model.

\subsection{Datasets and Evaluation Metrics}

\textbf{Datasets:}
Three widely used RGB-D saliency detection benchmark datasets are used to evaluate our \emph{cm}SalGAN method, i.e., NJUD~\cite{ju2014depth}, NLPR~\cite{peng2014rgbd} and STEREO~\cite{niu2012leveraging}.
NJUD dataset~\cite{ju2014depth} contains 2003 stereo images that are collected from the Internet, 3D movies and photographs acquired by a stereo camera. Then, the optical flow technique~\cite{sun2010secrets} is adopted to recover the depth maps.
NLPR dataset~\cite{peng2014rgbd} consists of 1000 images which are all taken by Kinect under different lighting conditions including both indoors and outdoors. The images in this dataset are selected from 5000 natural images and their depth maps and the saliency regions are annotated by five participants.
STEREO dataset~\cite{niu2012leveraging} consists of 797 stereo images.

For fair comparison, we adopt the same protocol to separate each dataset, as introduced in work~\cite{han2017cnns}.
The training subset contains 1400 and 650 samples from NJUD and NLPR dataset respectively. The validation set contains 100 samples from the NJUD and 50 samples from the NLPR dataset. The remaining samples in NJUD and NLPR and all the images in the STEREO dataset are used for testing.

\textbf{Evaluation Metrics:}
To achieve a comprehensive evaluation of the proposed \emph{cm}SalGAN saliency detection method and other comparison detectors,  we adopt four standard evaluation metrics to evaluate the predicted saliency maps, including Precision-Recall (PR) curve, S-measure scores, Mean Absolute Error (MAE) and maximal F-measure.
Specifically, Precision-recall curve~\cite{borji2015salient} is one of the most popular evaluation metric for saliency detection.
S-measure~\cite{fan2017structure} is a metric which evaluates both region-aware $S_{region}$  and object-aware $S_{object}$ structural similarity between saliency output and ground truth map, which can be formulated as:
\begin{equation}
\mathrm{S\textrm{-}measure} = \alpha S_{object} + (1 - \alpha) S_{region}
\end{equation}
MAE~\cite{perazzi2012saliency} is defined as the average pixel-wise absolute difference between the saliency output and the ground truth map which is defined as
\begin{equation}
MAE=\frac{1}{W \times H}\sum_{x=1}^{W}\sum_{y=1}^{H}\left | S\left ( x,y \right )-G\left ( x,y \right ) \right |
\end{equation}
The F-measure~\cite{achanta2009frequency} is a balanced mean of average precision and average recall which is calculated as
\begin{equation}
F =\frac{\left ( 1+\beta ^{2} \right )\times Precision \times Recall}{\beta ^{2}\times Precision \times Recall}
\end{equation}
where $\beta^{2}$ is set as 0.3, as suggested in most previous works~\cite{wang2019quality, pan2017salgan}. The definition of precision and recall are:
\begin{equation}
\label{Precision}
Precision = \frac{TP}{TP + FP}; 	~~~ Recall = \frac{TP}{TP + FN}
\end{equation}
where TP, FP, TN and FN denote the numbers of true positives, false positives, true negatives and false negatives, respectively.

\subsection{Implementation Details}

Due to the limited images in the RGB-D saliency detection datasets, similar to previous related works~\cite{zhao2019contrast, chen2019multi, han2017cnns, chen2019three, chen2018progressively}, we adopt horizontal flip and random crop to augment the training data.
In addition, we also add supervision with a smaller weight to each layer of the decoders so that being able to guide the decoder layers to learn better feature representation. The weight ratios of the five layers are set to $\{0.5, 0.5, 0.5, 0.8, 0.8\}$ respectively. In the cmSalGAN framework, the generator and discriminator are alternately trained according to batchsize. When training the discriminator, only the parameters of the discriminator are updated. When the generator is trained, the gradient of the discriminator is reversely transmitted to the generator and update the parameters of the generator.
It is also worthy to note that the proposed Adversarial Feature Learning module is only used in the training phase to consistently learn the multi-modal adversarial features. That is, only the two-stream generator and feature fusion module are used in the test phase. Our model is implemented based on PyTorch and all experiments are implemented with a NVIDIA 1080Ti GPU. The Adam \cite{kingma2014adam} optimizer is used for the training and the learning rate and batch size are set to 0.0001 and 4 respectively. We train the network for 180 epochs which cost 91 hours and each image takes 0.88s during the test phase. We fine-tune 10 epochs on the edge detection dataset~\cite{liu2019simple} when adding edge module. Our \emph{cm}SalGAN is an end-to-end network, which has no pre-training stage or other post-processing operations.

\begin{figure*}[htb]
\centering
\includegraphics[width=\textwidth]{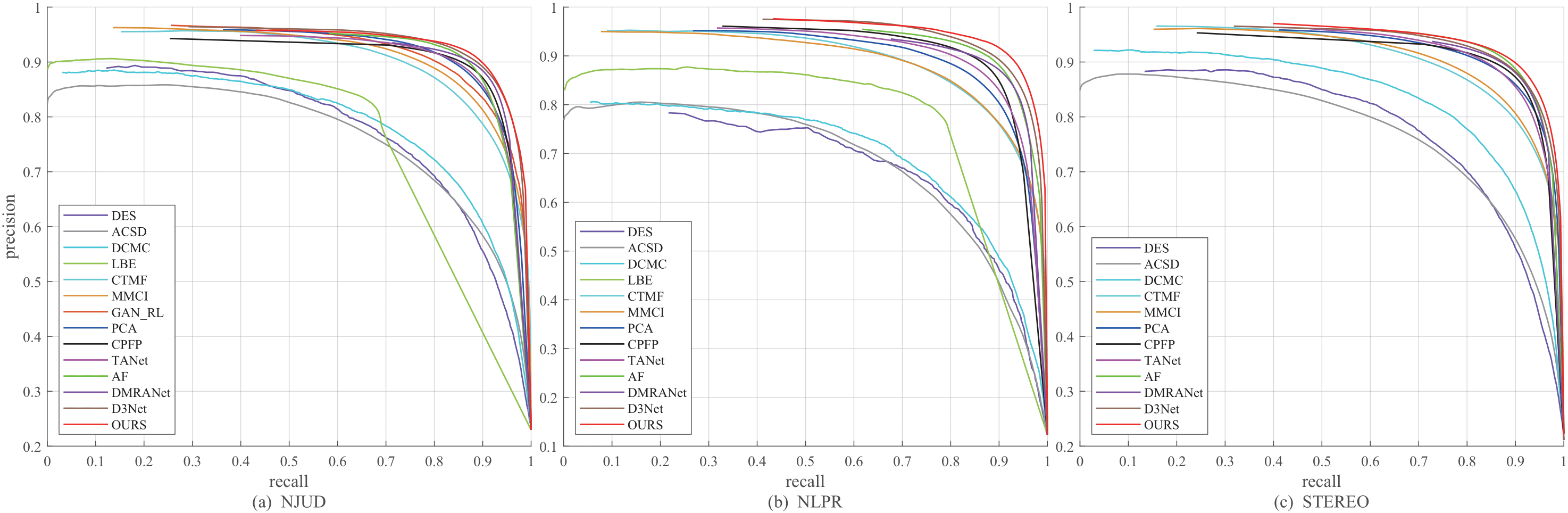}
\caption{Visual comparison P-R curves on NJUD, NLPR and STEREO respectively. }
\label{fig:PR}
\end{figure*}

\subsection{Comparison results}

\begin{table*}[htbp]
\caption{Comparison of different methods on NJUD, NLPR and STEREO datasets respectively. The evaluation measurements contain maximal F-measure, S-measure and MAE, respectively. The top 2 detection results are highlighted in red and green, respectively.}
\centering
\begin{tabular}{l|c|c|c|c|c|c|c|c|c}
\hline
\hline
& \multicolumn{3}{c|}{NJUD}& \multicolumn{3}{c|}{NLPR}& \multicolumn{3}{c}{STEREO}\\ \hline
Method &F-measure&S-measure& MAE          &F-measure&S-measure& MAE          &F-measure&S-measure& MAE      \\ \hline
DES\cite{cheng2014depth}   &0.7083 &0.6620 &0.2896   &0.6391 &0.5727 &0.3164   &0.7275 &0.6697 &0.2821   \\
ACSD\cite{ju2014depth}     &0.7110 &0.7018 &0.1905   &0.6436 &0.6987 &0.1560   &0.7171 &0.7138 &0.1840   \\
DCMC\cite{cong2016saliency}&0.7220 &0.6895 &0.1674   &0.6556 &0.7288 &0.1122   &0.7586 &0.7376 &0.1500   \\
LBE\cite{feng2016local}    &0.7456 &0.7003 &0.1490   &0.7576 &0.7769 &0.0731   &-      &-      &-        \\
CTMF\cite{han2017cnns}     &0.8441 &0.8490 &0.0847   &0.8255 &0.8599 &0.0561   &0.8385 &0.8529 &0.0867   \\
GAN-RL \cite{wang2019quality}&0.8679 &0.8681 &0.0803  &-      &-      &-    &-      &-      &-     \\
PCA\cite{chen2018progressively}&0.8722&0.8770&0.0591 &0.8410 &0.8736 &0.0437   &0.8700 &0.8800 &0.0606   \\
MMCI\cite{chen2019multi}   &0.8518 &0.8581 &0.0790   &0.8148 &0.8557 &0.0591   &0.8425 &0.8559 &0.0796   \\
TANet\cite{chen2019three}  &0.8737 &0.8782 &0.0605   &0.8632 &0.8861 &0.0410   &0.8705 &0.8775 &0.0591   \\
CPFP\cite{zhao2019contrast}&0.8767 &0.8777 &0.0533   &0.8675 &0.8884 &0.0360   &0.8738 &0.8792 &0.0514   \\
AF\cite{wang2019adaptive}  &0.8819 &0.8813 &0.0532   &0.8851 &0.9011 &0.0329   &0.8907 &0.8921 &{\color[HTML]{FE0000} 0.0472} \\
DMRANet\cite{piao2019depth}&0.8857 &0.8854 &0.0509   &0.8792 &0.8984 &0.0313   &0.8857 &0.8854 &{\color[HTML]{32CB00} 0.0475}   \\
D3Net\cite{fan2019rethinking}&0.8885&0.8946&0.0509   &0.8854 &0.9056 &0.0341   &0.8809 &0.8907 &0.0541   \\
GFNet\cite{LIU2020210}   & -     &0.8851 &0.0517   & -     &0.9075 &0.0296   & -     &0.8806 &0.0495   \\
\hline
\emph{cm}SalGAN-Edge & {\color[HTML]{32CB00} 0.8932} & {\color[HTML]{32CB00} 0.9000} & {\color[HTML]{32CB00} 0.0486}
       & {\color[HTML]{32CB00} 0.9037} & {\color[HTML]{32CB00} 0.9173} & {\color[HTML]{32CB00} 0.0286}
       & {\color[HTML]{32CB00} 0.8912} & {\color[HTML]{32CB00} 0.8978} & 0.0523 \\
\emph{cm}SalGAN & {\color[HTML]{FE0000} 0.8965} & {\color[HTML]{FE0000} 0.9034} & {\color[HTML]{FE0000} 0.0462}
       & {\color[HTML]{FE0000} 0.9070} & {\color[HTML]{FE0000} 0.9224} & {\color[HTML]{FE0000} 0.0267}
       & {\color[HTML]{FE0000} 0.8938} & {\color[HTML]{FE0000} 0.8999} & 0.0496 \\
\hline
\hline
\end{tabular}
\label{tab:TABEL}
\end{table*}

\begin{figure*}[htbp]
\centering
\includegraphics[width=\textwidth]{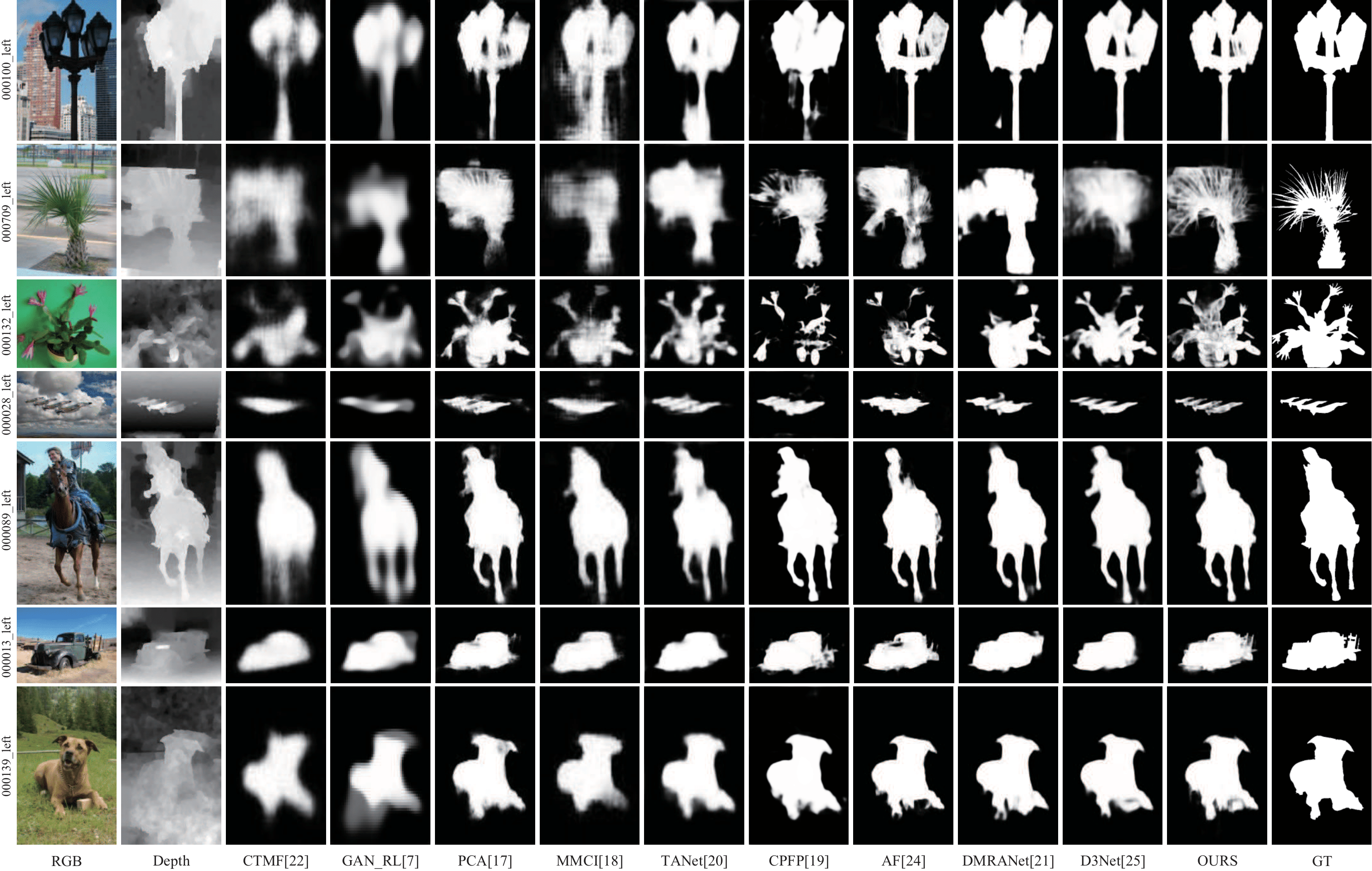}
\caption{Qualitative visual comparisons to other state-of-the-art CNNs-based methods and the last column GT represents the ground truth.}
\label{fig:compare}
\end{figure*}

We compare our \emph{cm}SalGAN with ten state-of-the-art deep learning based RGB-D salient object detection models including CTMF \cite{han2017cnns}, PCA \cite{chen2018progressively}, MMCI \cite{chen2019multi}, CPFP \cite{zhao2019contrast}, TANet \cite{chen2019three}, GAN-RL \cite{wang2019quality}, AF \cite{wang2019adaptive}, D3Net \cite{fan2019rethinking}, DMRANet\cite{piao2019depth} and GFNet \cite{LIU2020210}.
Note that CTMF \cite{han2017cnns}, PCA \cite{chen2018progressively}, MMCI \cite{chen2019multi} and TANet \cite{chen2019three} adopt VGG16 \cite{simonyan2014very} as backbone while D3Net \cite{fan2019rethinking} adopt resnet50 \cite{he2016deep}, and they all use cross-entropy loss function. GAN-RL \cite{wang2019quality} adopt VGG16 \cite{simonyan2014very} as backbone and use mean squared and adversarial loss, while GFNet \cite{LIU2020210} use cross-entropy and adversarial loss. CPFP \cite{zhao2019contrast} adopt VGG16 \cite{simonyan2014very} and use cross-entropy and contrast loss. AF \cite{wang2019adaptive} adopt VGG16 \cite{simonyan2014very} and use cross-entropy and edge-preserving loss. DMRANet\cite{piao2019depth} adopt VGG19 \cite{simonyan2014very} and use softmax-entropy loss.
In addition, we also report some traditional RGB-D salient object detection methods including DES \cite{cheng2014depth}, ACSD \cite{ju2014depth}, DCMC \cite{cong2016saliency} and LBE \cite{feng2016local}.
For comparison method CTMF \cite{han2017cnns}, PCA \cite{chen2018progressively}, MMCI \cite{chen2019multi}, TANet \cite{chen2019three}, ACSD \cite{ju2014depth}, DES \cite{cheng2014depth}, LBE \cite{feng2016local}, GAN-RL \cite{wang2019quality} and D3Net \cite{fan2019rethinking}, we directly evaluate the saliency maps of corresponding algorithm provided by the authors. {For GFNet \cite{LIU2020210}, we obtain the evaluation result directly from the paper.} For the other comparison methods, we run the original codes with default settings provided by the authors.
For evaluation measurements, we use the $evaluation\; tool \footnote{{ https://github.com/wenguanwang/SODsurvey/}}$ provided by Wang \emph{et al.} \cite{wang2019salient}.

Figure \ref{fig:PR} shows the comparison results on PR curve.
One can note that the proposed \emph{cm}SalGAN achieves the best performance on all three datasets, especially on the NLPR dataset. This suggests the effectiveness of the proposed  \emph{cm}SalGAN method.
Table \ref{tab:TABEL} summarizes the comparison results on MAE, maximal F-measure, and S-measure, respectively.
Our proposed method generally performs better than other comparison methods on most of the evaluation measurements and datasets. This also fully validates the effectiveness and advantages of the proposed \emph{cm}SalGAN method.
More specifically,
\emph{cm}SalGAN outperforms GAN-RL~\cite{wang2019quality} and GFNet~\cite{LIU2020210} which also employ adversarial feature learning for RGB-D saliency detection.
As discussed before, GAN-RL conducts adversarial learning for RGB and depth modality separately and adaptively fuses these results as post-processing by reinforcement learning architecture. GFNet construct a gated fusion module for cross-modal feature fusion through adversarial learning.
GAN-RL achieves 0.8679 on maximal F-measure on the NJUD dataset while the propose method achieves 0.8965,  GFNet achieves 0.8851 on S-measure on the NJUD dataset while the proposed method achieves 0.9034.
This obviously demonstrates the advantages of the proposed cross-view adversarial learning. In addition, as shown in the Table \ref{tab:TABEL} we can see that the variation \emph{cm}SalGAN-Edge (i.e. the framework without Edge module) also achieves the state-of-the-art performance.

\begin{table*}[htbp]
\centering
\caption{The evaluation of maximal F-measure, S-measure and MAE on NJUD, NLPR and STEREO respectively. $G_{Depth}$ and $G_{RGB}$ represent the depth and RGB branch of our generator respectively. $G_{RGB+Depth}$ represent fuse RGB and depth simply. '+G' and '+L' represent add the global and local attention scheme.}
\centering
\begin{tabular}{l|c|c|c|c|c|c|c|c|c}
\hline
\hline
       & \multicolumn{3}{c|}{NJUD}& \multicolumn{3}{c|}{NLPR}& \multicolumn{3}{c}{STEREO}\\ \hline
Method         &F-measure &S-measure &MAE         &F-measure &S-measure &MAE         &F-measure &S-measure &MAE      \\ \hline
$G_{Depth}$       &0.8103 &0.8351 &0.0948  &0.7884 &0.8276 &0.0739  &0.7119 &0.7659 &0.1289   \\
$G_{Depth}+LG$    &0.8239 &0.8498 &0.0834  &0.8308 &0.8622 &0.0566  &0.7283 &0.7754 &0.1196   \\
$G_{RGB}$         &0.8292 &0.8496 &0.0762  &0.8362 &0.8685 &0.0537  &0.8383 &0.8554 &0.0807   \\
$G_{RGB}+LG$      &0.8560 &0.8728 &0.0610  &0.8682 &0.8971 &0.0362  &0.8728 &0.8846 &0.0597   \\
$G_{RGB+Depth}$   &0.8680 &0.8811 &0.0599  &0.8708 &0.8959 &0.0403  &0.8593 &0.8757 &0.0668   \\
$G_{RGB+Depth}+G$ &0.8779 &0.8823 &0.0672  &0.8819 &0.9009 &0.0428  &0.8590 &0.8707 &0.0774   \\
$G_{RGB+Depth}+L$ &0.8736 &0.8843 &0.0600  &0.8775 &0.9028 &0.0374  &0.8578 &0.8724 &0.0663  \\
$G_{RGB+Depth}+LG$&0.8854 &0.8891 &0.0527  &0.8855 &0.9043 &0.0322  &0.8613 &0.8677 &0.0626  \\
\hline
\hline
\end{tabular}
\label{tab:RGB-Depth-tab}
\end{table*}

Figure \ref{fig:compare} shows some qualitative results to better demonstrate the advantages of the proposed saliency detection method. %
Intuitively, \emph{cm}SalGAN obtains the best saliency detection results compared with other approaches.
Specifically, our approach can produce more fine-grained details as highlighted in the salient region as shown in the $1^{st}$ and $2^{nd}$ rows in Figure \ref{fig:compare}. We can also observe that the depth image in the $7^{rd}$ row contains some misleading salience cues that make it difficult to distinguish interferences in depth information. However, our method still works well in such challenging scenarios. These conclusions can also be drawn from the rest of the images.

\subsection{Ablation Studies}

In this section, we conduct some ablation studies to better understand the effect of each component in our model. Specifically speaking, we will first check the effect of dual-modality fusion and the effect of global and local attention scheme in the generator. Then, we will discuss the effect of adversarial feature learning module and edge information. Finally, we will discuss the limitations of the proposed {RGB-D} salient object detection algorithm.

\textbf{Effectiveness of dual-modality fusion.}
We compare the proposed approach with only one modality used version to validate the effectiveness of the fusion module for RGB-D saliency detection. As shown in Table \ref{tab:RGB-Depth-tab}, it is easy to find that the result of fused RGB-D saliency detection is significantly better than only one modality used version on all the three datasets.
For example, the results of $G_{Depth}$ and $G_{RGB}$ achieve 0.8103 and 0.8292 on maximal F-measure on the NJUD dataset respectively, while the $G_{RGB+Depth}$ method which fuse RGB and depth features together achieves 0.8680 which is a significant improvement. Similar conclusions can also be drawn from other evaluation metrics and benchmark datasets. These experiments all validated the effectiveness of dual-modality fusion for saliency detection.

\begin{figure}[htbp]
\centering
\includegraphics[width=0.49\textwidth]{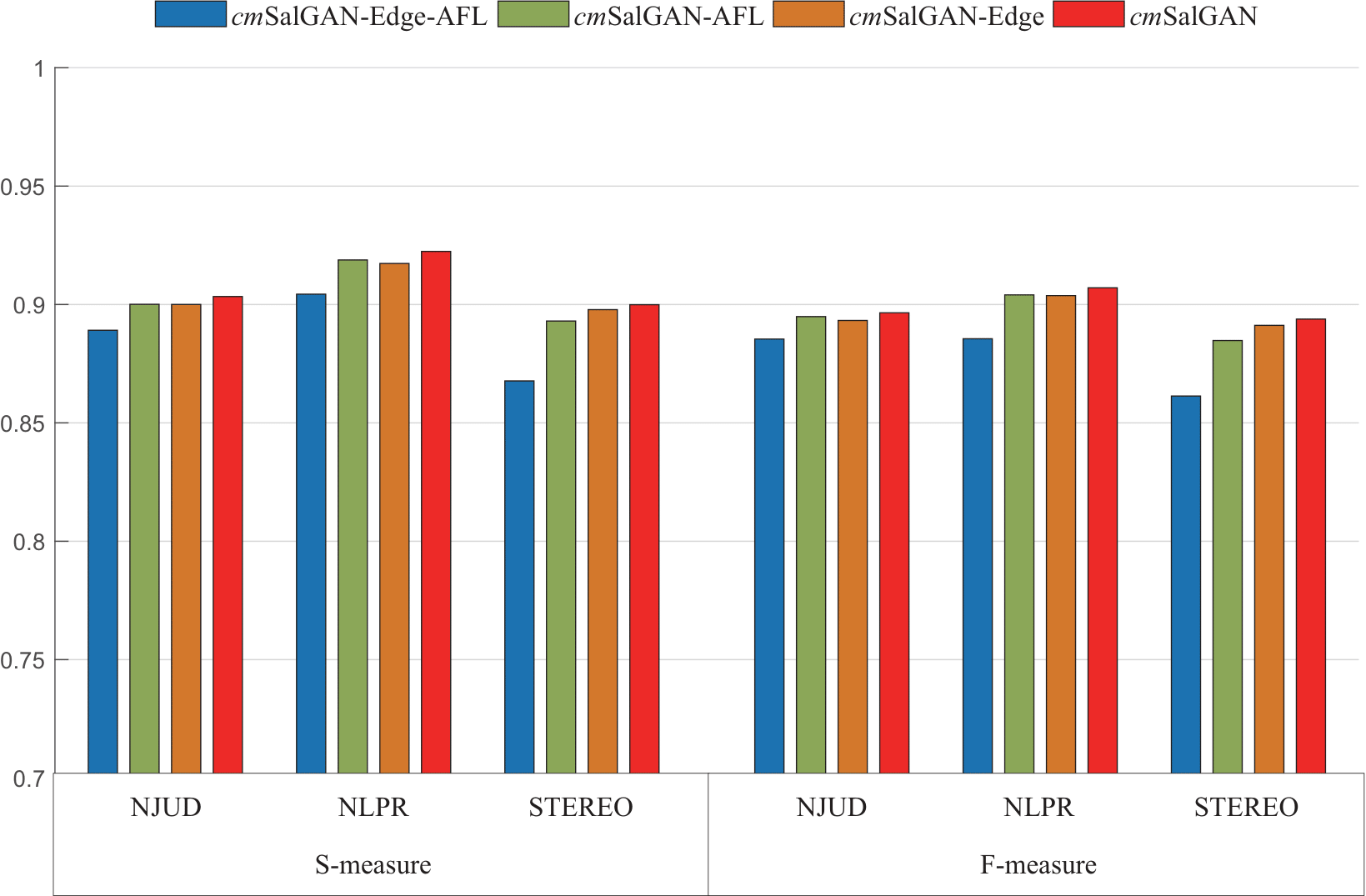}
\caption{Comparisons of S-measure and maximal F-measure to evaluate the contribution of AFL module and Edge module. '-Edge' and '-AFL' represent the \emph{cm}SalGAN network without Edge and AFL modules.}
\label{fig:ablation-bar}
\end{figure}

\textbf{Effectiveness of attention module.}
To validate the effectiveness of our used global or local attention modules, we conduct related analysis in the following subsection.
As reported in Table \ref{tab:RGB-Depth-tab}, all variants of our saliency detection algorithm can boost its performance with attention model on the used three benchmarks. More detail, the maximal F-measure score on the NJUD dataset of $G_{Depth}+LG$ and $G_{RGB}+LG$ all increased to 0.8239 and 0.8560. The $G_{RGB+Depth}$ achieves 0.8680 on maximal F-measure while $G_{RGB+Depth}+LG$ achieves 0.8854.

\textbf{Effectiveness of AFL and Edge components.}
To validate the effectiveness of AFL and Edge components in our network, we implement four variations of the model:

 1) \emph{cm}SalGAN-Edge-AFL that removes Edge and AFL module, only use BCE loss function in the training phase.

 2) \emph{cm}SalGAN-AFL that removes AFL module.

 3) \emph{cm}SalGAN-Edge that removes Edge module.

 4) \emph{cm}SalGAN that adds Edge and AFL module.

As shown in Fig. \ref{fig:ablation-bar}, we can note that:
(1) Remove the Edge and AFL module, the saliency detection results dropped in all three benchmark datasets.
(2) The overall performance can be  improved with the Edge module (i.e. the framework with BCE loss function and Edge module).
(3) Add AFL module could improve the performance of our saliency detection network, which validate the effectiveness of the AFL module. We can obtain 0.8932, 0.9037 and 0.8912 on the maximal F-measure on the NJUD, NLPR, and STEREO as shown in Table \ref{tab:TABEL}. And it achieves the state of the art performance.
(4) We introduce an edge detection module in the shallow layer of the encoder which can further improve the performance of our \emph{cm}SalGAN network. This experiment validates the effectiveness of the Edge module. When integrating the edge feature into our model, we can achieve better saliency detection performance. More detail, we can obtain 0.8965, 0.9070, 0.8938 on the maximal F-measure on the NJUD, NLPR and STEREO.

\begin{figure}[htbp]
\centering
\includegraphics[width=0.5\textwidth]{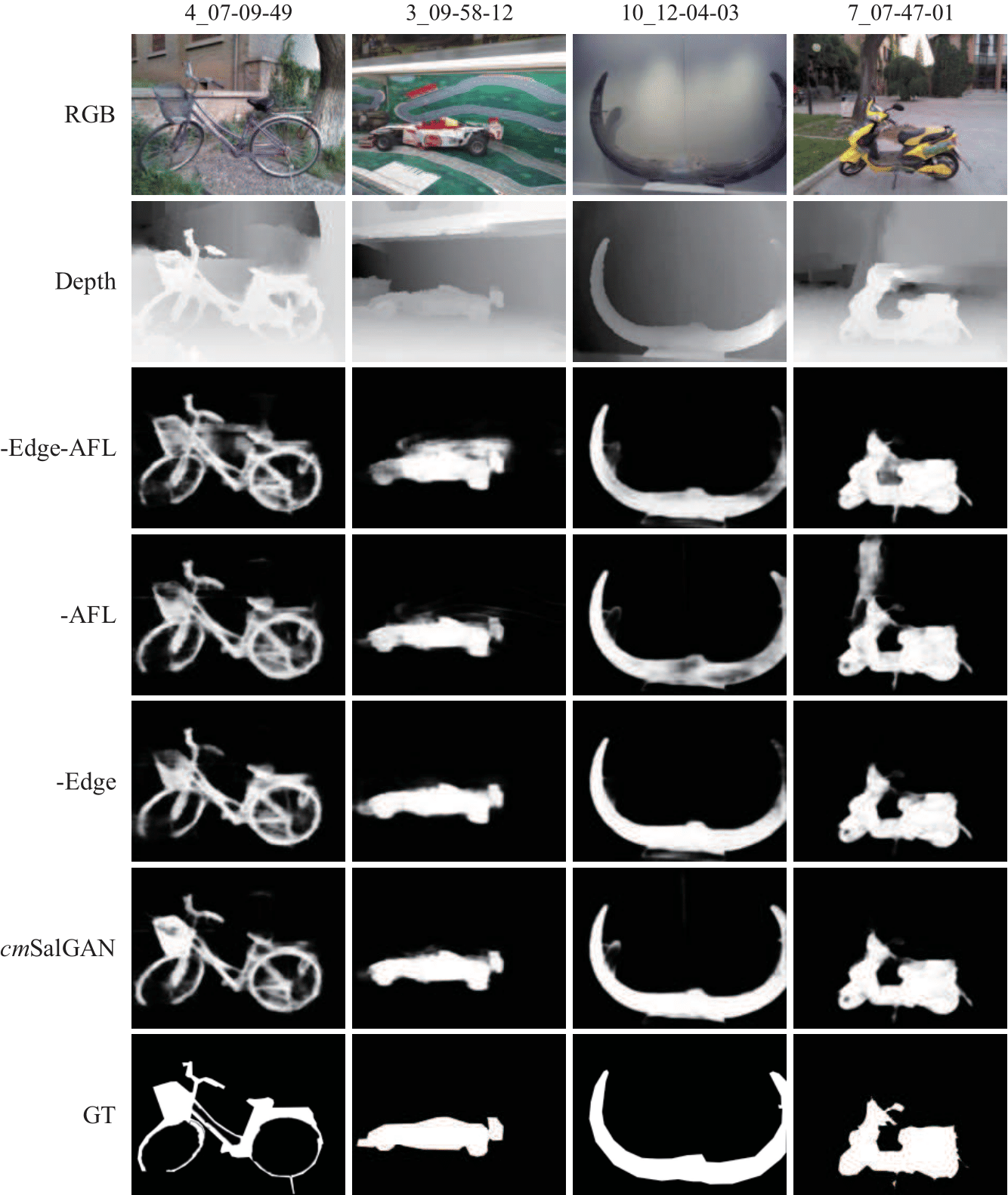}
\caption{Visualization of saliency detection results for different variations of the model. '-Edge' and '-AFL' represent the \emph{cm}SalGAN network without Edge and AFL modules.}
\label{fig:ablation-maps}
\end{figure}

We also conduct a visual comparison of saliency maps generated by these four variants of our model in Figure \ref{fig:ablation-maps}. From both qualitative and quantitative analysis, we can observe that the proposed adversarial feature learning module can significantly improve the deep representation learning for RGB-D saliency detection. And integrate the edge feature into our model could further improve the performance of the saliency detection network.

\section{Conclusion}
In this paper, we design an end-to-end RGB-D salient object detection framework based on generative adversarial feature learning. Our framework utilizes a two-stream generator to learn the feature representations of RGB and depth, respectively. We creatively fuse the features of RGB and depth with a feature embedding module to handle the limitations of single modality. More importantly, we conduct the adversarial feature learning between both RGB and depth modalities to boost their deep representations in the training stage. Our experiments validated the effectiveness of the proposed method on multiple saliency detection benchmarks. Note that, the proposed \emph{cm}SalGAN provides a general framework for RGB-D saliency detection tasks. Our module can be incorporated into many RGB-D algorithms to further improve their final performance.

\bibliographystyle{unsrt}
\bibliography{reference}

\begin{IEEEbiography}[{\includegraphics[width=1in,height=1.25in,clip,keepaspectratio]{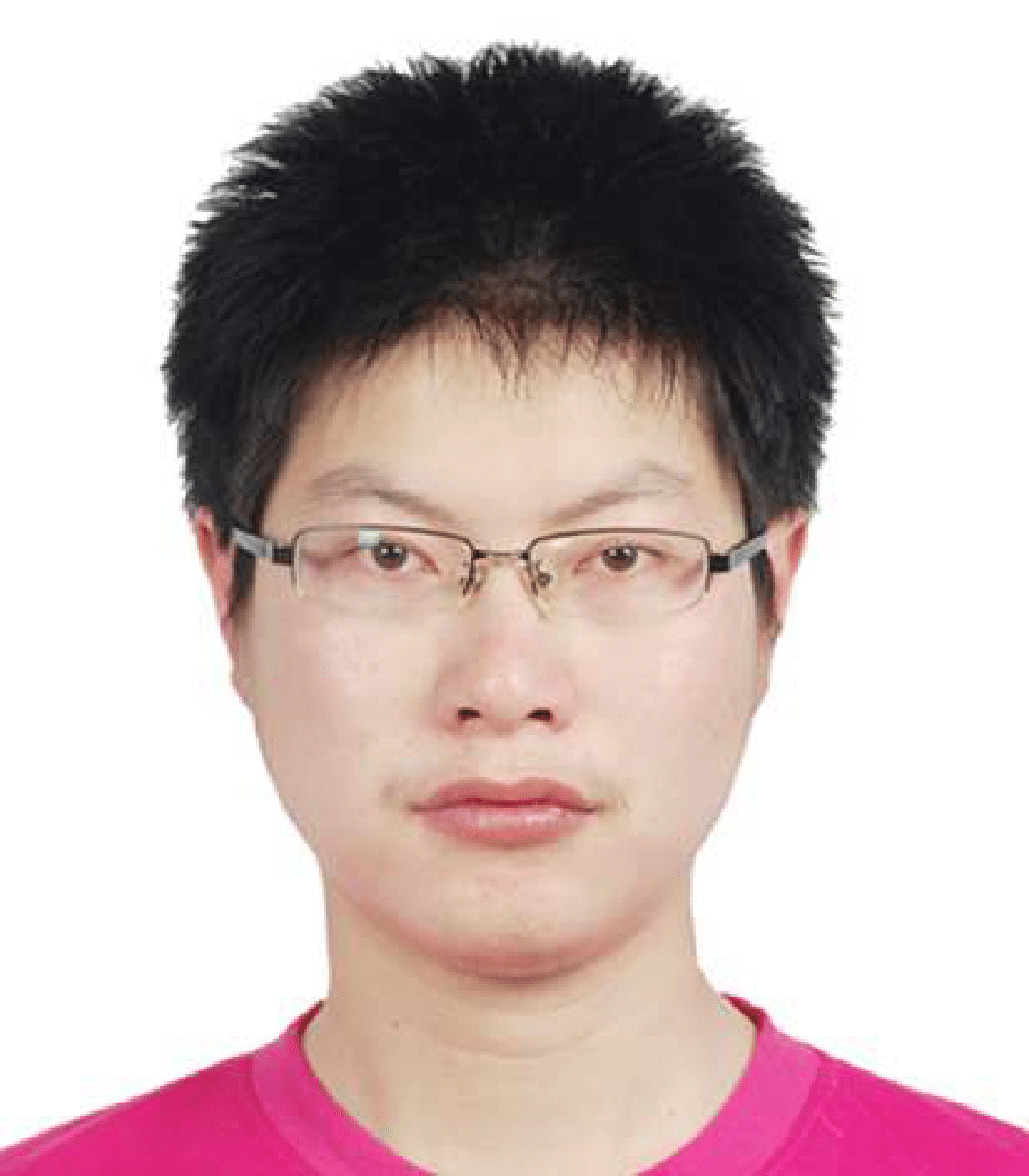}}]{Bo Jiang}
received the B.S. degrees in mathematics and applied mathematics and the  M.Eng.  and Ph.D. degree in computer science from Anhui University of China in 2009, 2012, and 2015, respectively. He is currently an associate professor in computer science at Anhui University. His current research interests include image feature extraction and matching, data representation and learning.
\end{IEEEbiography}

\begin{IEEEbiography}[{\includegraphics[width=1in,height=1.25in,clip,keepaspectratio]{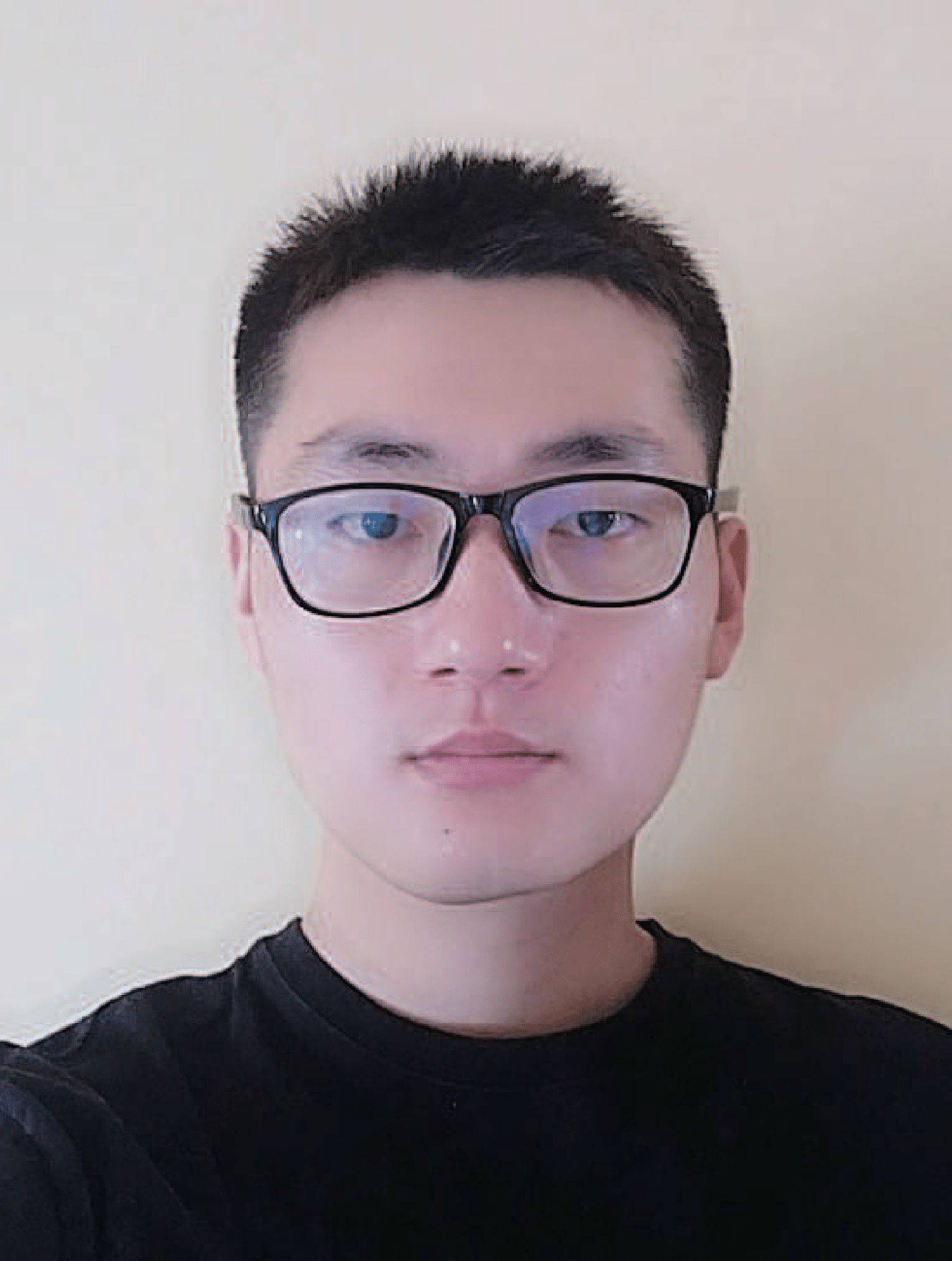}}]{Zitai Zhou}
is currently a Master student in computer science at Anhui University. His current research interests include saliency detection, RGB-D image analysis.
\end{IEEEbiography}

\begin{IEEEbiography}[{\includegraphics[width=1in,height=1.25in,clip,keepaspectratio]{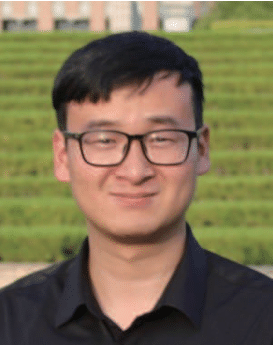}}]{Xiao Wang}
received the B.S. degree in West Anhui University, Luan, China, in 2013. He received the Ph.D. degree in computer science in Anhui University, Hefei, China, in 2019. He is now a postdoc in Pengcheng Laboratory. From 2015 and 2016, he was a visiting student with the School of Data and Computer Science, Sun Yat-sen University, Guangzhou, China. He also have a visiting at UBTECH Sydney Artificial Intelligence Centre, the Faculty of Engineering, the University of Sydney, in 2019. His current research interests mainly about computer vision, machine learning, pattern recognition and deep learning. He serves as a reviewer for a number of journals and conferences such as IEEE TCSVT, TIP, PR, CVPR, ICCV, ECCV and AAAI.
\end{IEEEbiography}

\begin{IEEEbiography}[{\includegraphics[width=1in,height=1.25in,clip,keepaspectratio]{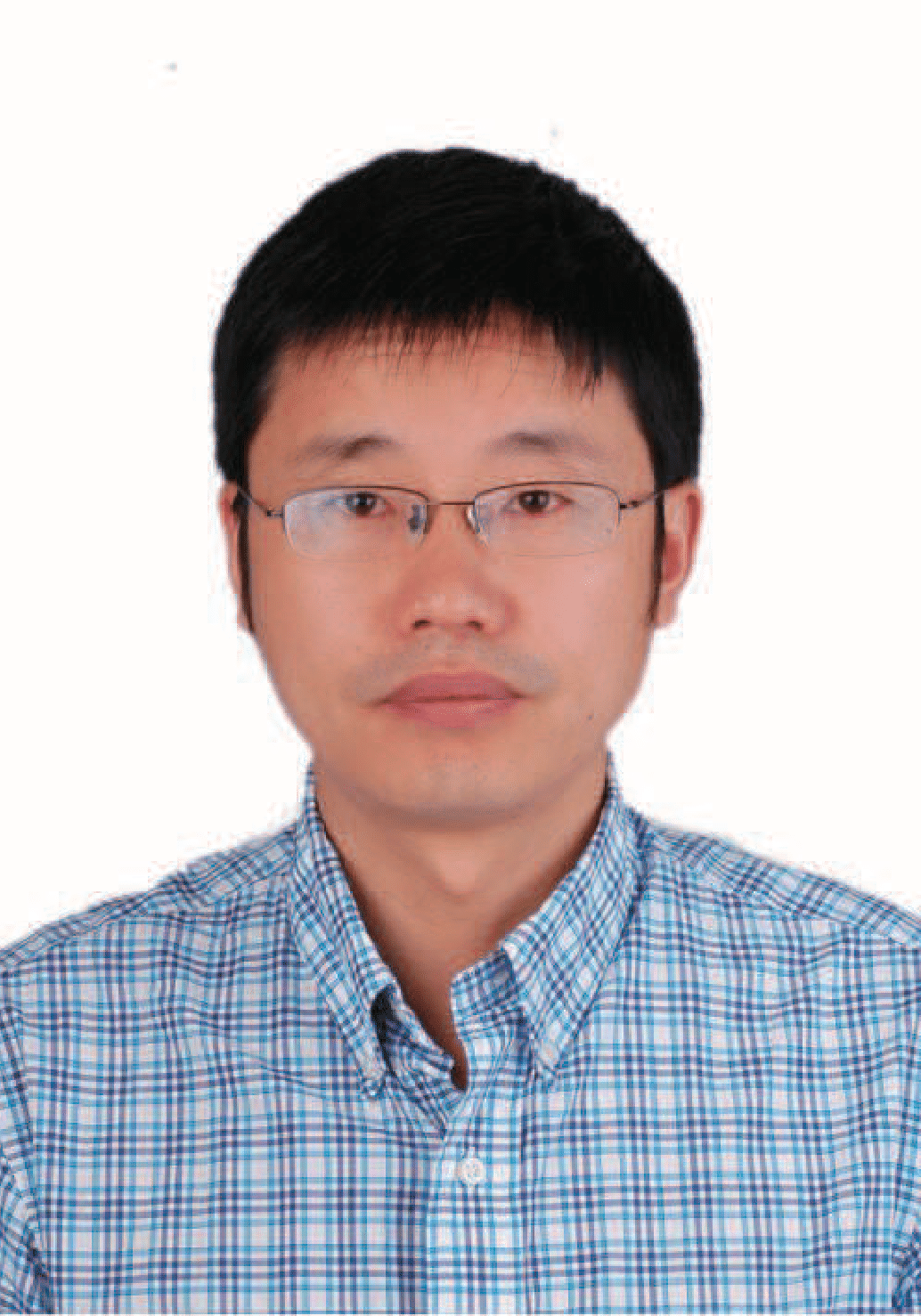}}]{Jin Tang}
received the B.Eng. degree in automation in 1999, and the Ph.D. degree in computer science in 2007 from Anhui University, Hefei, China. Since 2012, he has been a professor at the School of Computer Science and Technology at the Anhui University. His research interests include image processing, pattern recognition and computer vision.
\end{IEEEbiography}

\begin{IEEEbiography}[{\includegraphics[width=1in,height=1.25in,clip,keepaspectratio]{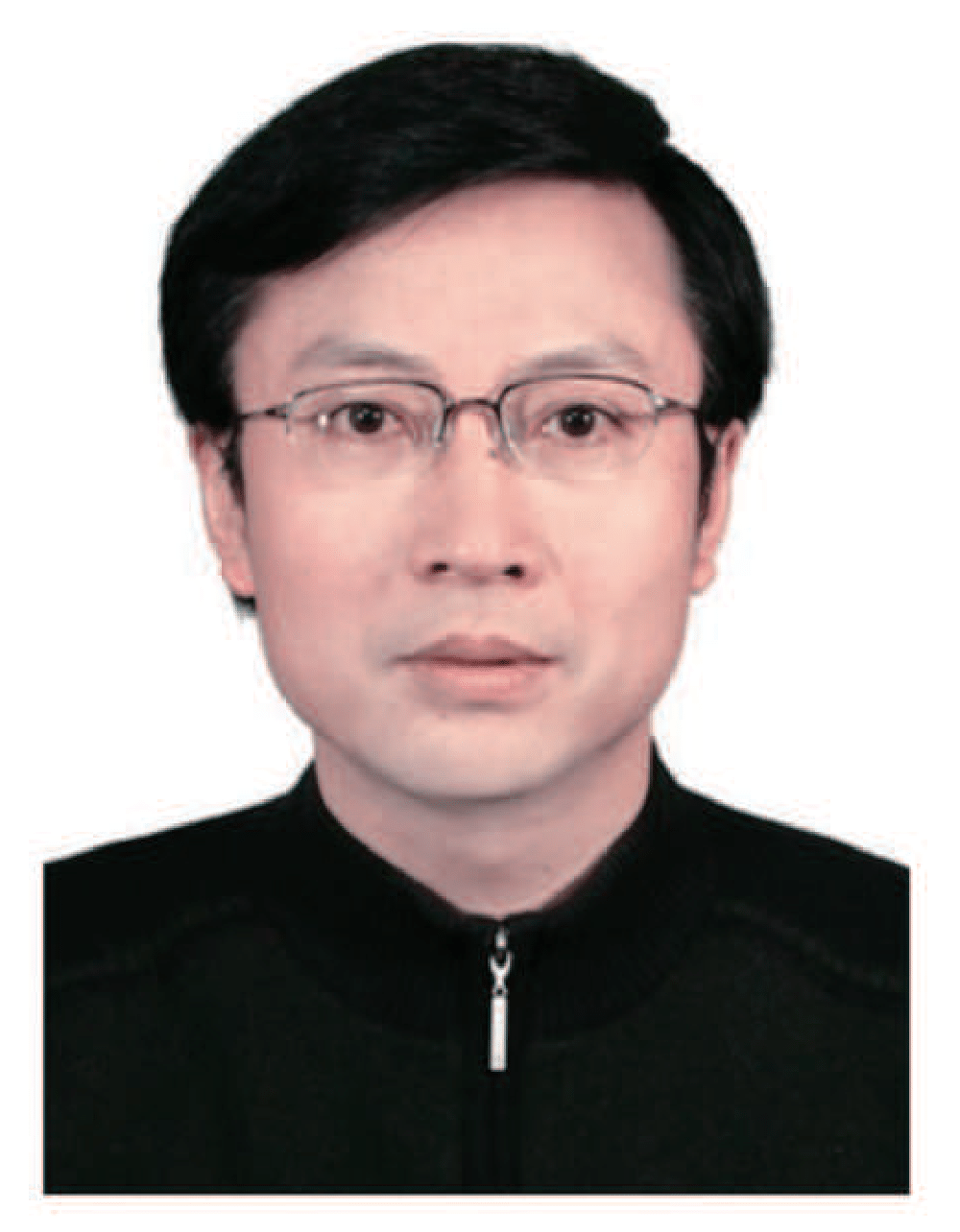}}]{Bin Luo}
received his B.Eng. degree in electronics and M.Eng. degree in computer science from Anhui University of China in 1984 and 1991, respectively. In 2002, he was awarded the Ph.D. degree in Computer Science from the University of York, the United Kingdom. He has published more than 200 papers in journal and refereed conferences. He is a professor at Anhui University of China. At present, he chairs the IEEE Hefei Subsection. He served as a peer reviewer of international academic journals such as IEEE Trans. on PAMI, Pattern Recognition, Pattern Recognition Letters, etc. His current research interests include random graph based pattern recognition, image and graph matching, spectral analysis.
\end{IEEEbiography}

\end{document}